\documentclass{article}

\PassOptionsToPackage{numbers, compress}{natbib}

    \usepackage[final]{neurips_2023}

\usepackage[utf8]{inputenc} 
\usepackage[T1]{fontenc}    
\usepackage{hyperref}       
\usepackage{url}            
\usepackage{booktabs}       
\usepackage{amsfonts}       
\usepackage{nicefrac}       
\usepackage{microtype}      
\usepackage{xcolor}         

\title{ReConTab: Regularized Contrastive Representation Learning for Tabular Data}

\usepackage[inline]{enumitem}
\usepackage{times}

\usepackage{bm}
\usepackage{tikz}
\usepackage{scalerel}
\usepackage{xspace}
\usepackage{pifont}
\usepackage{multirow}
\usepackage{tikz}
\usepackage{comment}
\usepackage{amsmath,amssymb} 
\usepackage{color}
\usepackage{float}
\usepackage{booktabs}       
\usepackage{indentfirst}
\usepackage{multirow}
\usepackage{makecell}
\usepackage{pifont}
\usepackage{everyshi}
\newcommand\doubleplus{+\kern-1.3ex+\kern0.8ex}
\newcommand{\cmark}{\ding{51}}%
\usepackage{colortbl}
\usepackage{url}            
\usepackage{booktabs}       
\usepackage{amsfonts}       
\usepackage{nicefrac}       
\usepackage{microtype}      
\usepackage{xcolor}         
\usepackage{makecell}
\usepackage{multirow}
\usepackage{amssymb}
\usepackage{amsmath}
\usepackage{algorithm,algcompatible}
\usepackage{caption}
\usepackage{subcaption}
\usepackage{placeins}

\author{%
  Suiyao Chen \thanks{These authors contributed equally to this work.} \\
  University of South Florida\\
  Tampa, FL 33620\\
  \texttt{suiyaochen@usf.edu} \\
  \And
  Jing Wu $^{\ast}$\\
  University of Illinois at Urbana-Champaign \\
  Champaign, IL 61820\\
  \texttt{jingwu6@illinois.edu} \\
  \AND
  Naira Hovakimyan \\
  University of Illinois at Urbana-Champaign  \\
   Champaign, IL 61820\\
  \texttt{nhovakim@illinois.edu} \\
  \And
  Handong Yao \\
  University of Georgia \\
  Athens, GA 30602 \\
  \texttt{handong.yao@uga.edu} \\
}

\begin{document}

\maketitle
\graphicspath{{./figures/}}

\begin{abstract}%
Representation learning stands as one of the critical machine learning techniques across various domains. Through the acquisition of high-quality features, pre-trained embeddings significantly reduce input space redundancy, benefiting downstream pattern recognition tasks such as classification, regression, or detection. Nonetheless, in the domain of tabular data, feature engineering and selection still heavily rely on manual intervention, leading to time-consuming processes and necessitating domain expertise. In response to this challenge, we introduce ReConTab, a deep automatic representation learning framework with regularized contrastive learning. Agnostic to any type of modeling task, ReConTab constructs an asymmetric autoencoder based on the same raw features from model inputs, producing low-dimensional representative embeddings. Specifically, regularization techniques are applied for raw feature selection. Meanwhile, ReConTab leverages contrastive learning to distill the most pertinent information for downstream tasks. Experiments conducted on extensive real-world datasets substantiate the framework's capacity to yield substantial and robust performance improvements. Furthermore, we empirically demonstrate that pre-trained embeddings can seamlessly integrate as easily adaptable features, enhancing the performance of various traditional methods such as XGBoost and Random Forest.
\end{abstract}


\section{Introduction}

In the last decade, representation learning has made remarkable strides in fields like computer vision and natural language processing, revolutionizing the way we extract valuable insights from image and text data. However, several critical industries, including healthcare\citep{qayyum2020secure,chen2017personalized,chen2019claims}, manufacturing\citep{borisov2022deep, chen2017multi,bingjie2023optimal,chen2020some}, agriculture\citep{liakos2018machine, wu2022optimizing, tao2022optimizing} and various engineering fields\citep{xu2019importance, wang2019inverse,chen2020optimal,Che_Liu_Li_Huang_Hu_2023}, still heavily rely on structured tabular data. Researchers traditionally leverage domain expertise for feature selection\citep{covert2019deep}, model refinement\citep{wang2018surrogate,wang2017sensitivity} and uncertainty quantification\citep{chen2018data,wang2023inverse,wang2019gaussian}. With high-quality hand-crafted features, it is typically believed that traditional tree-based models are able to automatically capture the feature importance and interactions, without additional tuning. 

While manual feature engineering has been effective, it comes with challenges. The process of crafting high-quality features for tabular data is labor-intensive and lacks the guarantee of optimal performance. Feature selection often requires iterative experimentation, making it resource and time-intensive. In tackle these challenges, recent research endeavors have sought to harness the potential of deep representation learning for more efficient feature engineering in tabular data.

However, tabular data presents unique hurdles that have hindered its integration with the remarkable success of deep learning in other domains. In contrast to text data, where tokens are inherently discrete, or images, where pixels exhibit spatial correlations, tabular data encompasses a diverse mix of continuous, categorical, and ordinal values. These values can exhibit complex interdependencies and correlations, adding layers of complexity to the modeling process. Moreover, unlike the structured nature of images or the sequential nature of text, tabular data lacks inherent positional information to capture the intrinsic meanings or learn explicit representations. 

In this paper, we proposed ReConTab, a transformer-based framework to automatically generate high-quality embeddings as features for classification model improvement. Our framework consists of an asymmetric autoencoder (AE) architecture, which is able to extract the most critical information from raw features to provide substantial performance improvement and robustness for downstream classification tasks. Moreover, ReConTab can be effectively trained in both self- and semi-supervised modes. This adaptability ensures the model to perform well across 
various training scenarios, irrespective of the availability of labeled data. The contributions are summarized as follows: 

\begin{itemize}[noitemsep,topsep=0pt]

    \item[$\bullet$] We proposed a transformer-based automatic feature engineering framework, which is agnostic to modeling tasks, with scalability and adaptability.

    \item[$\bullet$] We designed a novel AE architecture with regularization and contrastive learning for an enhanced feature learning process. 

    \item[$\bullet$] We conducted a comprehensive empirical study on various public datasets that demonstrates the superiority of the proposed work in performance lift and robustness.

    \item[$\bullet$] We demonstrated that representative embeddings extracted from raw features can serve as readily applicable features, seamlessly augmenting the performance of various conventional classification models such as logistic regression and tree-based models, etc.

\end{itemize}

\section{Related Work}
\subsection{Classical Models} Various traditional machine-learning methods have been developed for tabular data classification and regression tasks. When it comes to modeling linear relationships, Logistic Regression (LR) \cite{wright1995logistic} and Generalized Linear Models (GLM) \citep{hastie2017generalized} are the prominent choices. For those seeking tree-based models, Decision Trees (DT) \citep{breiman2017classification} are popular options. Additionally, there are various ensemble methods based on DT, such as XGBoost \citep{chen2016xgboost}, Random Forest \citep{breiman2001random}, CatBoost \citep{prokhorenkova2018catboost}, and LightGBM \citep{ke2017lightgbm}. These ensemble methods are widely embraced in the industry for their ability to model complex non-linear relationships, enhance interpretability, and handle various feature types, including null values and categorical features.

\subsection{Deep Learning Models} 
The current research landscape has a prominent trend focusing on applying deep learning techniques to tabular data. This movement has given rise to diverse neural architectures, each of which is designed to enhance performance within tabular data domain. These architectures can be broadly classified into several categories \citep{borisov2022deep, gorishniy2021revisiting}.
Firstly, there are supervised methods that harness the power of neural networks, including well-known models like ResNet \citep{he2016deep}, SNN \citep{klambauer2017self}, AutoInt \citep{song2019autoint}, and DCN-V2 \citep{wang2021dcn}, to improve the handling of tabular data.
Secondly, there exist hybrid approaches that seamlessly integrate decision trees with neural networks, resulting in end-to-end training. This category includes innovative techniques like NODE \citep{popov2019neural}, GrowNet \citep{badirli2020gradient}, TabNN \citep{ke2018tabnn}, and DeepGBM \citep{ke2019deepgbm}.
Thirdly, transformer-based methods have emerged, allowing models to learn from attention-spanning features and data points. Notable examples in this class include TabNet \citep{arik2021tabnet}, TabTransformer \citep{huang2020tabtransformer}, and FT-Transformer \citep{gorishniy2021revisiting}.
Lastly, representation learning methods are gaining prominence, emphasizing effective information extraction through self- and semi-supervised learning techniques. Noteworthy models in this realm encompass VIME \citep{yoon2020vime}, SCARF \citep{bahri2021scarf}, and SAINT \citep{somepalli2021saint}. These approaches align seamlessly with the growing emphasis on representation learning in the field.

\subsection{Self- and Semi-supervised Representation Learning} 
In computer vision, deep representation learning methodologies have emerged as potent tools, capitalizing on self- and semi-supervised training paradigms \citep{kolesnikov2019revisiting, ericsson2022self, li2015semi}. These methodologies exhibit a dichotomy, falling into two distinct categories of innovation.
The first category of deep representation learning methods is rooted in generative models, particularly autoencoders \citep{kingma2013auto}. A striking exemplar within this genre is the Masked AutoEncoder (MAE) architecture introduced by \cite{he2022masked}. MAE features an asymmetric encoder-decoder architecture purposefully crafted for the extraction of embeddings from images. Impressively, the framework demonstrates the capability to capture spatiotemporal information \citep{feichtenhofer2022masked} and extends seamlessly to various domains such as 3D space \citep{jiang2022masked} and multiple scales \citep{reed2022scale}. Notably, akin masking strategies, prevalent in the Natural Language Processing (NLP) community \citep{devlin2018bert}, have also been transposed into the tabular data landscape \citep{arik2021tabnet, huang2020tabtransformer, yin2020tabert}. Furthermore, VIME \citep{yoon2020vime} presents a method reminiscent of MAE in the tabular data context. VIME perturbs and encodes each data sample within the feature space through the involvement of two estimators. Subsequently, these estimators use decoders to reconstruct both a binary mask and the original, uncorrupted data samples, demonstrating versatility in information extraction.

The second category predominantly revolves around the contrastive learning paradigm and strategically employs data augmentation techniques. Within this domain, prominent models harnessed momentum-update strategies \citep{he2020momentum, chen2020improved,wu2023genco,wu2023hallucination}, embraced the concept of large batch sizes \citep{chen2020simple}, incorporated stop-gradient operations \citep{chen2021exploring}, spatiotemporal information\cite{wu2023extended}, or even introduced an online network tasked with predicting the output of a target network \citep{grill2020bootstrap}. Notably, these concepts, initially designed for image data, have gracefully transcended into the arena of tabular data. An exemplar of such adaptation is found in SCARF \citep{bahri2021scarf}, which ingeniously incorporates the principles of SimCLR \citep{chen2020simple} to pre-train the encoder. This pre-training procedure employs a subset of feature corruption as a pivotal data augmentation method. Furthermore, the work of \cite{somepalli2021saint} exemplifies a contrastive framework tailored to tabular data, introducing SAINT, computing both column- and row-wise attentions.

\subsection{Regularization}
Regularization techniques, pivotal in machine learning and statistical modeling, mitigate overfitting and enhance generalization by introducing penalty terms into the loss function. Early approaches such as Ridge Regression, which applies L2 regularization to linear models \citep{cortes2012l2}, and Lasso Regression \citep{meier2008group}, which implements L1 regularization, paved the way for modern regularization methods. The Elastic Net \citep{zou2005regularization} combines these approaches to strike a balance between feature selection and coefficient shrinkage, while Dropout \citep{srivastava2014dropout} and Batch Normalization \citep{ioffe2015batch} cater specifically to neural networks, fostering robust and generalized representations. Other techniques like early stopping \citep{prechelt2002early} and weight decay \citep{loshchilov2017decoupled} further complement the regularization arsenal. Bayesian approaches introduce probabilistic frameworks, such as Bayesian regression \citep{bishop2003bayesian} and Gaussian Processes \citep{rasmussen2003gaussian}, integrating prior beliefs and data likelihood. Recent trends encompass adversarial training \citep{miyato2018virtual} to enhance model robustness and graph regularization techniques \citep{feng2019graph} for graph-based data modeling tasks. As machine learning continues to advance, regularization remains vital for model generalization and robustness.

\section{Method}
In this section, we present ReConTab, our comprehensive approach for tabular data self- and semi-supervised representation learning. First, we outline the process of the regularization method. Second, we formulate the feature corruption process. The self-supervised training process is illustrated in the third sub-section, without knowing the task labels. The fourth sub-section elucidates our novel semi-supervised training method, wherein we leverage labels for contrastive learning. Finally, we expound on our utilization of pre-trained encoders and embeddings to improve downstream tasks.

\subsection{Regularization}
\label{sec: regularization}
We apply regularization \citep{covert2019deep, wu2023contrastive} on the input layer by introducing a penalty term $\lambda \|\boldsymbol{W}\|_p$ into the loss function, where $\boldsymbol{W}$ represents the input weights, $\lambda$ is the regularization parameter and $p$ is the specific norm for the penalty. The idea behind is to prevent similar features from weighing too much in loss objective and to learn more robust representation, especially when highly correlated features are present. For example, if we can reconstruct features A, B, and C with only feature A, then B and C should be assigned less weights.

\subsection{Feature Corruption}
\label{sec: corruption}
It's common for the generative-based representation approach to use data augmentation techniques to generate robust feature embeddings. One of the most promising approaches is feature corruption, which has also been used in this paper to enhance our model's performance. Considering the original dataset $\mathcal{X} \subseteq \mathbb{R}^M$, given any tabular data point $x_{i}$, we have its $j$-th feature as $x_{i_j}$, where $x_{i} = (x_{i_1}, x_{i_2},..., x_{i_M}), j \subseteq M$, with $M$ representing the dimension of features and $i$ denoting the sample index. In our approach, for each sample, we stochastically select $t$ features from the pool of $M$ features and replace them with corrupted features denoted as $c$. To elaborate, we generate $c$ from the distribution $\widehat{\mathcal{X}}{i_j}$, where $\widehat{\mathcal{X}}{i_j}$ represents the uniform distribution over $\mathcal{X}_{i_j} = \left\{ x_{i_j}: x_{i} \in \mathcal{X}\right\}$.

\subsection{Self-supervised Learning}
\label{sec: ssl}
Self-supervised learning of ReConTab aims to learn informative representations from unlabeled data (Algorithm~\ref{algo:self-supervised}). For each of the two data samples, $x_1$ and $x_2$, we apply input weights and add feature corruption to obtain corrupted data. Then we encode the corrupted data using an encoder, $f$, resulting in two features, $z_1$ and $z_2$. The decoder $d$ will decode the learned embeddings to reconstruct $\hat{x}_1$ and $\hat{x}_2$ respectively, from where we can define the reconstruction loss $\mathcal{L}_\text{reconstruction}$ for two samples $x^{1}$ and $x^{2}$ as the mean squared error (MSE) between input features and reconstructions, shown as:
\begin{align}
     \mathcal{L}_\text{reconstruction} = \frac{1}{M}\sum_{j=1}^{M} (x_{1_j} - \hat{x}_{1_j})^2 + \frac{1}{M}\sum_{j=1}^{M} (x_{2_j} - \hat{x}_{2_j})^2.
\end{align}
\begin{algorithm}
\small
\begin{algorithmic}
\REQUIRE unlabeled data  $\mathcal{X} \subseteq \mathbb{R}^M$, batch size $B$, encoder $f$, decoder $d$, mean squared error {(MSE)}, input weights $\boldsymbol{W} \subseteq \mathbb{R}^M$, regularization parameter $\lambda$ and specific norm for penalty $p$. 

\FOR {two sampled mini-batch $ \left\{x_{i}^{1}, y_{i}^{1}\right\}_{i=1}^{B} \subseteq \left\{\mathcal{X},\mathcal{Y}\right\}$ and $ \left\{x_{i}^{2},y_{i}^{2}\right\}_{i=1}^{B} \subseteq \left\{\mathcal{X},\mathcal{Y}\right\}$} 
    \STATE for each sample $x_{i}^{1}$ and $x_{i}^{2}$, 

    \STATE apply input weights
    $x_{i}^{1} = x_{i}^{1}\boldsymbol{W}$,
    $x_{i}^{2} = x_{i}^{2}\boldsymbol{W}$, for $i \in [B]$

    \STATE apply feature corruption, define the corrupted feature as: 
    $\breve{x}_{i}^{1}$ and $\breve{x}_{i}^{2}$, for $i \in [B]$
    
    \STATE data encoding:  
    
    $z_{i}^{1}=f(\breve{x}_{i}^{1})$, $z_{i}^{2}=f(\breve{x}_{i}^{2})$, for $i \in [B]$
    
    \STATE data reconstruction:
    
    $\hat{x}^{1}_{i} = d(z_{i}^{1})$, 
    $\hat{x}^{2}_{i} = d(z_{i}^{2})$, for $i \in [B]$

    \STATE define reconstruction loss $\mathcal{L}_\text{reconstruction}=$ MSE$({x}^{1}_{i}, \hat{x}^{1}_{i})+$MSE$({x}^{2}_{i}, \hat{x}^{2}_{i})$

    \STATE define penalty as $\lambda \|\boldsymbol{W}\|_p$

    \STATE update encoder $f$ and decoder $d$ to minimize $\mathcal{L}_\text{reconstruction}$ and $\lambda \|\boldsymbol{W}\|_p$ using RMSProp.
    \ENDFOR
    \end{algorithmic}
    \caption{Self-supervised learning}
    \label{algo:self-supervised}
\end{algorithm}

Therefore, the loss function for self-supervised learning can be defined as:
\begin{align}
        \mathcal{L}_\text{self} =  \mathcal{L}_\text{reconstruction} + \lambda \|\boldsymbol{W}\|_p,
\end{align}

\subsection{Semi-supervised Learning}
\label{sec: semi}
\begin{figure}[H]
	\centering
	\includegraphics[width=1\textwidth]{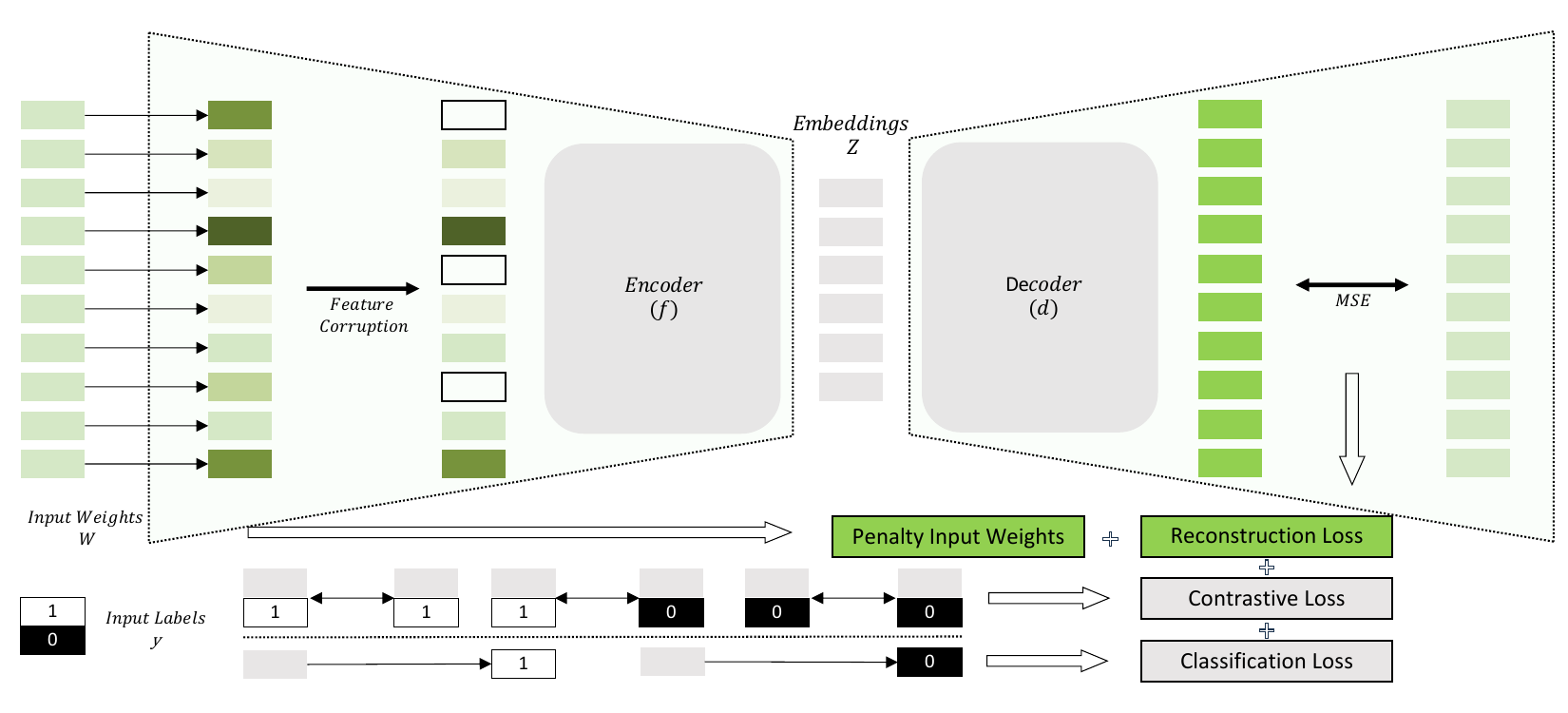}
	\caption{\small Proposed AE architecture with contrastive loss and input weights regularization}
	\label{ReConAE}
\end{figure}

We further improve the pre-training process through semi-supervised learning to take advantage of labeled data, as shown in Figure~\ref{ReConAE}. In self-supervised learning, we only compute the MSE between reconstructed data and original data as the reconstruction loss $ \mathcal{L}_\text{reconstruction}$. With labels introduced, we can pose additional constraints to the encoded embeddings $z_1$ and $z_2$. One is for label prediction to compute the prediction loss (illustrated by classification loss $\mathcal{L}_\text{classification}$ through the context). To be specific, $z_1$ and $z_2$ are fed to the same multi-layer perceptron (MLP) that maps from the embedding space to the label space. We can also define the cross-entropy loss for classification task as:
\begin{align}
    \mathcal{L}_\text{classification} = -\left( y_{1} \log(\hat{y}_{1}) + y_{2} \log(\hat{y}_{2}) \right),
\end{align}
where $\hat{y}_{1}$ and $\hat{y}_{2}$ are predicted labels computing a MLP, i.e., $\hat{y}_{1}= \text{MLP}(z_1)$ and $\hat{y}_{2} = \text{MLP}(z_2)$.

\subsubsection{Contrastive Loss}
\label{sec: contrast}
We further introduce the contrastive loss $\mathcal{L}_{\text{contrastive}}$ in the loss function by forming contrastive pairs ($z_1$, $z_2$) of embeddings in the bottleneck layer with respect to the classification labels ($y_1$, $y_2$). With this constraint, the model is enforced to maximize the similarity between embeddings with the same label and minimize the similarity between embeddings with different labels, thus capturing the discriminative features for the classification labels and better aligning with downstream tasks. Algorithm~\ref{algo:Contrastive_loss} formally defines the contrastive loss in the proposed model, which is a variation from the original contrastive learning \citep{hadsell2006dimensionality} and relevant to these extensions \citep{chen2020simple,tao2022supervised,gharibshah2022local}.
\begin{algorithm}
\small
	\begin{algorithmic}
\REQUIRE data embeddings $\mathcal{Z}$ from unlabeled data $\mathcal{X} \subseteq \mathbb{R}^M$, binary labels $\mathcal{Y} \subseteq \mathbb{R}$, batch size $B$, encoder $f$, decoder $d$, contrastive loss margin $m$, distance function $D(\cdot)$
 
\FOR {two sampled mini-batch $ \left\{z_{i}^{1}, y_{i}^{1}\right\}_{i=1}^{B} \subseteq \left\{\mathcal{X},\mathcal{Y}\right\}$ and $ \left\{z_{i}^{2},y_{i}^{2}\right\}_{i=1}^{B} \subseteq \left\{\mathcal{X},\mathcal{Y}\right\}$} 
    \STATE for each sample embedding $z_{i}^{1}$ and $z_{i}^{2}$, 

    \STATE define contrastive loss:
    \FOR{${i=1}$ to $B$}
		\IF{$y^1_i$ = $y^2_i$}
		\STATE $y^{12}_i=1$ for the pair $(z^1_i, z^2_i)$ \quad \text{// $z^1_i$ is deemed similar to $z^2_i$}
		\ELSE{} 
		\STATE $y^{12}_i=0$ for the pair $(z^1_i, z^2_i)$ \quad \text{// $z^1_i$ is deemed dissimilar to $z^2_i$}
		\ENDIF
		\STATE $d_i = D(z^1_i, z^2_i)$ \quad \text{// calculate the distance of two embeddings in the pair}
		\STATE $c_i = (y^{12}_i)\frac{1}{2}d_i^2 + (1-y^{12}_i)\frac{1}{2}\max(0,m-d_i)^2 $ \quad \text{// calculate the contrastive loss of the pair}
    \ENDFOR
        \STATE $\mathcal{L}_{contrastive}= \frac{1}{B}\sum c_i$

        \STATE update encoder $f$ and decoder $d$ to minimize $\mathcal{L}_{contrastive}$ using RMSProp.
    \ENDFOR
    \end{algorithmic}
    \caption{Contrastive Loss for Semi-supervised learning}
    \label{algo:Contrastive_loss}
\end{algorithm}

During the optimization stage, we combine the two additional losses with the self-supervised learning loss $\mathcal{L}_\text{self}$ and define the semi-supervised learning loss function $\mathcal{L}_\text{semi}$ as follows:
\begin{align}
        \mathcal{L}_\text{semi} =  \mathcal{L}_\text{self} + \alpha * \mathcal{L}_\text{classification} + \beta * \mathcal{L}_\text{contrastive},
\end{align}
where $\alpha$ and $\beta$ are used to seek balance among multiple losses and set to 1 as default, respectively.

\subsection{Downstream Fine-Tuning}
Drawing inspiration from established representation learning paradigms \citep{he2020momentum, chen2020improved, chen2020simple, bahri2021scarf}, we embrace an end-to-end fine-tuning strategy for the pre-trained encoder $f$ from ReConTab, utilizing the complete labeled dataset. This approach entails the seamless integration of the encoder with an additional linear layer, thereby granting the flexibility to unlock and adapt all its parameters to align with the specific requirements of downstream supervised tasks. Additionally, we can harness the potential of the salient feature $s$ as a versatile plug-and-play embedding. Through the fusion of $z$ with its original counterpart $x$, we construct enriched data points. This innovative approach serves to amplify inherent data characteristics, thereby assisting in the establishment of distinct decision boundaries. As a result, we anticipate notable enhancements in classification tasks when employing the concatenated features as the input for conventional models like Random Forest or LightGBM.

\section{Experiments and Results}
In this section, we present the results of our extensive experiments conducted on diverse public datasets to highlight the effectiveness of our proposed method, ReConTab. This section is structured into two parts for clarity and comprehensiveness. In the first part, we provide essential details regarding the experiments. This includes information about the public datasets for experiments, the preprocessing steps applied to these datasets, the architecture of our models, and specific training procedures. This transparency ensures the reproducibility of our findings.

\begin{figure}[H]
	\centering
	\includegraphics[width=0.6\textwidth]{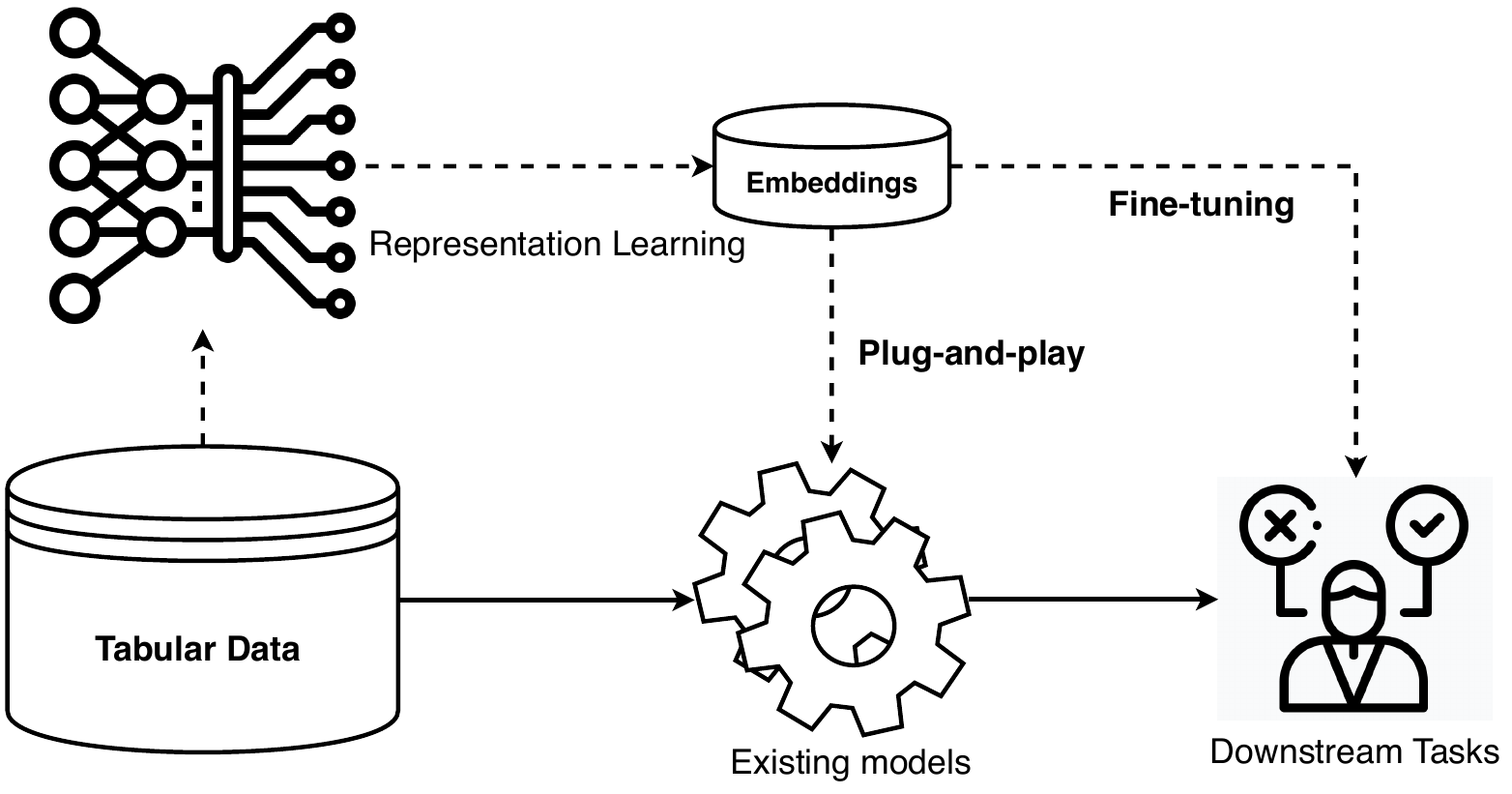}
	\caption{\small Illustration of usages of pre-trained encoders and embeddings. 1) The first option could be to fine-tune the pre-trained encoder directly for downstream tasks. This option usually achieves the optimal results but needs additional computation. 2) The second option is to concatenate the pre-trained embeddings with the original datasets, which requires no additional training and computation but still benefits the downstream tasks with considerable improvements in evaluation metrics.}
   \label{ReConTab_plugNplay}
\end{figure}

In the second part, we assess the performance of ReConTab through various empirical studies. We conduct a thorough comparison between ReConTab and mainstream deep learning methods as well as traditional methods. Meanwhile, we showcase the versatility of ReConTab using it as an automatic feature engineering tool. Specifically, we demonstrate how ReConTab can enhance the performance of traditional models such as XGBoost, Random Forest, and LightGBM by seamlessly integrating its salient features as plug-and-play embeddings, as shown in Figure~\ref{ReConTab_plugNplay}. This strategy simplifies the feature engineering process and eliminates additional complexity in traditional models training.

\subsection{Preliminaries for Experiments}

\subsubsection{Public Datasets} 

We evaluate the performance of ReConTab on a standard benchmark from \cite{somepalli2021saint}, including Bank (BK) \citep{moro2014data}, Blastchar (BC) \citep{ibm2019}, Arrhythmia (AT) \citep{liu2008isolation}, Arcene (AR) \citep{asuncion2007uci}, Shoppers (SH) \citep{sakar2019real}, Volkert (VO) \citep{automlchallenges} and MNIST (MN) \citep{xiao2017fashion}. Five of the datasets focus on binary classification, and two of them focus on multi-class classification tasks. Importantly, the datasets employed in our experiments exhibit significant diversity. They encompass a wide range of characteristics, including varying sample sizes, ranging from 200 to 495,141 samples, and feature dimensions spanning from 8 to 784, encompassing both categorical and numerical features. Among these datasets, some exhibit missing data, while others are complete, and there is a mix of well-balanced datasets as well as those presenting highly skewed class distributions. This diversity allows us to comprehensively evaluate the performance and robustness of our proposed approach across a spectrum of real-world data scenarios.

\subsubsection{Preprocessing of Datasets}
To handle categorical features, we employ a backward difference encoder as described in \cite{potdar2017comparative}. Addressing the issue of missing data, we take a two-step approach. Initially, we remove any features that lack values across all samples. Subsequently, for the remaining missing values, we apply distinct imputation strategies based on the feature type. Numerical features are imputed using the mean value, while categorical features are filled with the most frequent category observed within the dataset. Moreover, we ensure data uniformity by employing a min-max scaler for dataset scaling. In cases involving image-based data, we flatten the images into vectors, treating them akin to tabular data. This approach aligns with established practices found in prior works such as \cite{yoon2020vime} and \cite{somepalli2021saint}.

\subsubsection{Model Architectures} 
The ReConTab model architecture features a transformer-based shared network with three layers and two attention heads. This architecture is tailored for processing input data with a dimensionality determined by the shape of the training dataset. Additionally, the decoder remains a one-layer network with a sigmoid activation function. In the downstream fine-tuning stage, we add a linear layer after the encoder $f$ to accommodate classification or regression tasks as needed.

\subsubsection{Training Details} 
The ReConTab model is trained with a batch size of 128 over 1000 epochs, employing a learning rate of 0.0001. Gaussian masking is applied to the input data with a masking ratio of 0.3. The model's output dimension is set to half of the input data dimension. A contrastive loss with a margin of 2 is used during training, along with L2 normalization. Additionally, a regularization coefficient of 0.01 is applied to introduce a penalty term based on the L2 norm of the standard deviation of the Gaussian mask. During training, data is divided into two batches, and various loss components, including feature reconstruction loss, classification loss, contrastive loss, and regularization penalty, are computed to guide the optimization process. These training configurations ensure effective representation learning while controlling model behavior.

\subsubsection{Metrics} 
Given that the majority of the tasks in our analysis involve binary classification, we employ the AUROC (Area Under the Receiver Operating Characteristic curve) as our primary metric for assessing performance. AUROC effectively quantifies the model's ability to distinguish between the two classes in the dataset. However, for the two multi-class datasets, VO and MN, we utilize accuracy on the test set as the metric for comparing performance.

\subsection{Results on the Benchmarks}
We show performance comparisons using chosen datasets and present the summarized results in Table~\ref{tab: further}. These results encompass evaluations employing both traditional models and more recent deep-learning techniques. In the majority of cases, ReConTab exhibits remarkable improvements, outperforming all baseline methods and reaffirming its superiority across a range of datasets and scenarios. 
However, it is important to note that, on BK, SH, and VO datasets, ReConTab achieved suboptimal results when compared to the best models. This observation aligns with previous research conclusions that the tabular domain presents unique challenges, with no single method universally excelling \citep{gorishniy2021revisiting}. Nonetheless, ReConTab still gives the best performance over all of the deep-learning-based models and the second-best results over all of the methods. Meanwhile, this outcome warrants further investigation to uncover the specific factors contributing to this variation in performance.

\begin{table*}
    \centering
    \resizebox{1\textwidth}{!}{
    \begin{tabular}{l c>{\columncolor[gray]{0.95}}c>{\columncolor[gray]{0.8}}c | c>{\columncolor[gray]{0.95}}c>{\columncolor[gray]{0.8}}c |c>{\columncolor[gray]{0.95}}c>{\columncolor[gray]{0.8}}c |c>{\columncolor[gray]{0.95}}c>{\columncolor[gray]{0.8}}c |c>{\columncolor[gray]{0.95}}c>{\columncolor[gray]{0.8}}c |c>{\columncolor[gray]{0.95}}c>{\columncolor[gray]{0.8}}c |c>{\columncolor[gray]{0.95}}c>{\columncolor[gray]{0.8}}c  }
    \toprule
    \textbf{Dataset size}   & \multicolumn{3}{c}{45211} & \multicolumn{3}{c}{7043} & \multicolumn{3}{c}{452} & \multicolumn{3}{c}{200} & \multicolumn{3}{c}{12330} & \multicolumn{3}{c}{58310} & \multicolumn{3}{c}{518012} \\   
    \textbf{Feature size}   & \multicolumn{3}{c}{16} & \multicolumn{3}{c}{20} & \multicolumn{3}{c}{226} & \multicolumn{3}{c}{783} & \multicolumn{3}{c}{17} & \multicolumn{3}{c}{147} & \multicolumn{3}{c}{54}  \\   
    \midrule
    \textbf{Dataset} & \multicolumn{3}{c}{\textbf{BK}} & \multicolumn{3}{c}{\textbf{BC}} & \multicolumn{3}{c}{\textbf{AT}} & \multicolumn{3}{c}{\textbf{AR}} & \multicolumn{3}{c}{\textbf{SH}} & \multicolumn{3}{c}{\textbf{VO}$\bigstar$}  & \multicolumn{3}{c}{\textbf{MN}$\bigstar$} \\
    \midrule
    \textbf{Raw Feature ($x$)}  &\cmark &   &\cmark &\cmark &   &\cmark &\cmark &   &\cmark &\cmark &   &\cmark &\cmark &   &\cmark &\cmark &   &\cmark &\cmark &   &\cmark   \\
    \textbf{Distilled Feature ($s$)}  &  &\cmark &\cmark &  &\cmark &\cmark &  &\cmark &\cmark &  &\cmark &\cmark &  &\cmark &\cmark &  &\cmark &\cmark &  &\cmark &\cmark \\
    \midrule
    Logistic Reg.   & 0.907 &0.907 & 0.909      &0.892  &0.892 &0.895       & 0.862&0.864  &0.866     & \underline{0.916} &0.914 &0.918         & 0.870 &0.871 &0.873      & 0.539&0.540 &0.543        & 0.899 &0.902 &0.905     \\
    
    Random Forest   & 0.891 &0.892 & 0.894      &0.879  &0.880 &0.884       & 0.850&0.856  &0.861     & 0.809 &0.809 &0.811         & 0.929 &0.928& 0.930        
    &0.663  &0.665 &0.669       &0.938 &0.938 & 0.942 \\
    
    XGboost         & 0.929 &0.928 & 0.930      &0.906  &0.903 &0.906       & 0.870&0.871  &0.883     & 0.824 &0.822 &0.826      &0.925 &0.925&0.927              
    &\textbf{0.690}  &{0.690} &0.692       &{0.958} &0.959  &0.963 \\
    
    LightGBM        & \textbf{0.939} &0.933 & 0.939      &0.910  &0.909 &0.912       & 0.887&0.888  &0.907     & 0.821 &0.822 &0.825      &\textbf{0.932} &0.933&0.936  & 0.679 &0.680 &0.682       &0.952 &0.953 &0.954 \\
    
    CatBoost        & 0.925 &0.928 & 0.932      &\underline{0.912}  &0.910 &0.914       & 0.879&0.880  &0.889     & 0.825 &0.827 &0.833  &\underline{0.931} &0.932&0.935             &0.664  &0.665 &0.670       &0.956 &0.958 &0.968  \\
    
    MLP             & 0.915 &0.919 & 0.920      &0.892  &0.893 &0.898       & \underline{0.902}&0.904  &0.908     & 0.903 &0.904 &0.904          &0.887 &0.887&0.890         &0.631  &0.631 &0.636       &0.939 &0.940 &0.940 \\
     \midrule   
    VIME            & 0.766 & - & -    &0.510& - & -   & 0.653 & - & -        & 0.610 & - & -      &0.744 & - & -      &0.623 & - & -  &0.958  & - & -\\
    TabNet          & 0.918 & - & -    &0.796 & - & -   & 0.521 & - & -           &0.541  & - & -      &0.914 & - & -      &0.568 & - & -  &\textbf{0.968}  & - & - \\
    TabTransformer       & 0.913 & - & -    &0.817 & - & -   & 0.700 & - & -           &0.868  & - & -      &0.927 & - & -      &0.580 & - & - &0.887  & - & -\\
    \midrule
    ReConTab(Self-Sup.) &{0.908}   & - & -  &{0.898}   & - & -      &{0.873} & - & - & {0.887}  & - & - &{0.920}   & - & - &{0.619}   & - & - &{0.956}   & - & - \\
    
    \textbf{ReConTab(Semi-Sup.)} &\underline{0.929}   & - & -  &\textbf{0.913}   & - & -     &\textbf{0.907} & - & - & \textbf{0.918}  & - & - &\underline{0.931}   & - & - &\underline{0.680}   & - & - &\textbf{0.968}   & - & - \\

    \bottomrule
    \multicolumn{22}{l}{\large $``-"$ indicates the experiments are not applicable for the corresponding methods to demonstrate the benefits of plug-and-play embeddings.} \\
    \end{tabular}
    }
    \caption{\small Comparison of different methods on the classification tasks. For each method and dataset, we report three categories 1) raw features only, 2) salient features only, 3) plug-and-play features. The best results are shown in \textbf{Bold}, second-best results are \underline{Underlined}. Columns added with $\bigstar$ are multi-class classification tasks, reporting accuracy. The other results of binary classification tasks are evaluated with AUROC.}
  \vspace*{-3mm} 
    \label{tab: further}
\end{table*}

\subsection{Results as Plug-and-Play Embeddings}
As previously mentioned, ReConTab has learned features that can significantly impact the decision boundaries in classification tasks. In the plug-and-play setting from Figure~\ref{ReConTab_plugNplay}, our experimental results demonstrate the immense value of integrating these salient features with the original data as additional features. To be more specific, the performance of traditional models obtains relatively marginal improvement with only distilled features, as shown in the light gray columns of Table~\ref{tab: further}.  While the improvement is relatively modest, it aligns with our expectations. The potential absence of original information in this scenario results in a less substantial performance boost. Larger gains without fine-tuning come from the concatenation of original and distilled features. Notably, this integration enhances the performance of every method, leading to improvements in evaluation metrics (e.g., AUROC) across various datasets, as shown in the dark gray columns of Table~\ref{tab: further}. 


\subsection{Ablation Studies}

In this section dedicated to ablation studies, we delve into the crucial components of ReConTab, assessing the significance of the parameter, i.e., feature corruption rate. Our analysis encompasses all the datasets listed in Table~\ref{tab: further}, employing consistent data preprocessing and optimization strategies throughout the experiments. In Table~\ref{tab: ablation2}, we thoroughly examine the most advantageous feature corruption ratio. After extensive analysis, we find that the optimal corruption ratio is approximately 0.3. Therefore, we've adopted this value as the default for all previously reported experiments. However, it's important to emphasize that this chosen ratio may not always be the best fit for every dataset. Additionally, we've noticed interesting patterns in the datasets themselves. Datasets with more complex features, like VO or MN, tend to benefit from larger corruption ratios because they often contain redundant features. This observation aligns with previous research discussed in \cite{grinsztajn2022tree} regarding tabular data. On the flip side, for datasets with simpler, lower-dimensional features like BC, using smaller corruption ratios in our experiments might lead to better results.

\begin{table}
    \centering
    \resizebox{0.47\textwidth}{!}{
    \begin{tabular}{l c | c | c | c | c | c | c }
    \toprule
    \textbf{Ratio} & \multicolumn{1}{c}{\textbf{0.0}} & \multicolumn{1}{c}{\textbf{0.1}} & \multicolumn{1}{c}{\textbf{0.2}} & \multicolumn{1}{c}{\textbf{0.3}} & \multicolumn{1}{c}{\textbf{0.4}} & \multicolumn{1}{c}{\textbf{0.5}}  & \multicolumn{1}{c}{\textbf{0.6}} \\
    \midrule
    \textbf{BK}  & 0.918 &{0.920}  & 0.928 &\textbf{0.929}   & 0.922 &  {0.917}  & 0.881    \\
    \midrule
    \textbf{BC}  & 0.889 &{0.897}  & 0.906 &\textbf{0.913}   & 0.910 &  {0.901}  & 0.896    \\
    \midrule
    \textbf{AT} & 0.889  &{0.894}  & 0.901 &\textbf{0.905}   & 0.903 &  {0.890}  & 0.884   \\
    \midrule
    \textbf{AR} & 0.904  &{0.911}  & 0.913 &\textbf{0.918}   & 0.915 &  {0.909}  & 0.901   \\
    \midrule
    \textbf{SH} & 0.902  & {0.914}  & 0.924 & \textbf{0.931}   & {0.920} &  {0.909}  & 0.904    \\
    \midrule
    \textbf{\textbf{VO}$\bigstar$} & 0.667  &{0.674}  & 0.676 &{0.680}   & \textbf{0.681} &  {0.670}  & 0.663    \\
    \midrule
    \textbf{\textbf{MN}$\bigstar$} & 0.935  &{0.942}  & 0.951 &\textbf{0.959}   & \textbf{0.959} &  {0.941}  & 0.932    \\
    \bottomrule
    \end{tabular}
    }
    \caption{\small Ablation of corruption ratio. Columns added with $\bigstar$ are multi-class classification tasks, reporting their accuracy. The other results of binary classification tasks are evaluated with AUC.}
    \label{tab: ablation2}
    \vspace{-3mm} 
\end{table}

\section{Conclusion}
As we observe the evolution of potent representation learning techniques tailored for different types of data from computer vision and natural language processing, we embark on a journey to extend their remarkable performance into new domains, such as tabular data. Drawing inspiration from related endeavors that address this challenge from the vantage points of contrastive learning and generative modeling, we present ReConTab — an innovative self- and semi-supervised framework designed for representation learning and feature distillation. The features learned through ReConTab exhibit superior performance in downstream tasks, obviating the need for extensive exploration of hand-crafted features. Furthermore, these features manifest as discernible, low-dimensional representations that seamlessly enhance the capabilities of various traditional models. We hold a strong conviction that this research marks a pivotal milestone in the pursuit of more representative, efficient, and structured representations for tabular data. 


\bibliographystyle{plain}
\bibliography{nips2023}


\end{document}